# Generalised Object Detection and Semantic Analysis: Casino Example using Matlab

Othman Ahmad

Universiti Malaysia Sabah

## Abstract

*Matlab version 7.1 had been used to detect playing cards on a Casino table and the suits and ranks of these cards had been identified. The process gives an example of an application of computer vision to a problem where rectangular objects are to be detected and the information content of the objects are extracted out. In the case of playing cards, it is the suit and rank of each card. The image processing system is done in two passes. Pass 1 detects rectangular shapes and template matched with a template of the left and right edges of the cards. Pass 2 extracts the suit and rank of the cards by matching the top left portion of the card that contains both rank and suit information, with stored templates of ranks and suits of the playing cards using a series of if-then statements.*

## 1. Introduction

A Casino table had used to illustrate the techniques widely described in internet tutorials and textbooks to detect objects. The textbooks are listed in the references but the internet tutorials, which are not listed in the references, are just as important in order to implement the system for the recognition and semantic analysis of playing cards.

### 1.1. Software

The choice of development software is crucial. For this project, Mathlab version 7 was chosen. There are other alternatives such as Mathematica and Maple. The open software Octave is compatible with some versions of Matlab but it should be able to use Matlab Image Processing toolbox, which is used when this system was developed.

### 1.2. Casino Environment

This system is developed with a Casino environment in mind. It means that the images should be very clear because of the excellent lighting in Casinos. The playing cards are also well known because they are supplied by the Casinos. The number of samples taken is also limited to detection of playing cards but there is no reason why the system cannot be extended to include more samples in order to improve detection accuracy.

### 1.3. Occam's razor

In accordance with the principle of Occam's razor, the simplest algorithms had been used. For template matching, cross-correlation technique had been used instead of the more exotic neural networks, despite the many references in this paper, which refer to neural networks. In addition to that, only routines and techniques that are available in Matlab were used.

### 1.4. Modifications for other Objects

Despite the title of this paper, subsequent methodologies, which will be discussed, will refer to the detection of playing cards. In order to modify the system for other types of objects, the potential developer must master the writing of codes in Matlab first.
In order to learn about Matlab, the help and demo functions in Matlab are most useful. Just type "help" in the command prompt area of Matlab. Start by studying the functions used in this paper or the internet tutorials listed in the Appendix. Proceed with developing a more thorough understanding by reading through the textbooks on image processing as given by the references below[7][8][9][10].

## 2. Literature Review

### 2.1. Wen-Yuan Chen, Chin-Ho Chung

Wen-Yuan Chen, Chin-Ho Chung [1] claimed to have achieved "high veracity and highly robust recognition under a variety of conditions", for detecting playing cards. Experimental results show that this proposed scheme can exactly identify ranks and suits of the poker images 100% of the time, even when 40% noise is added, or when intensity level is increased or decreased 40%. It may be good for noisy environment but not necessary in typical casino environment. The method tests the entire card for playing card recognition requiring much processing and database storage.

## 2.2. Chunhui (Brenda) Zheng

In Chunhui (Brenda) Zheng [3], a rotational invariant template matching method is proposed in this paper to enable accurate playing card recognition. Some prior approaches for card recognitions have been made, but they were all unsuccessful. Character segmentation algorithms and affine transformation are then used in the playing card recognition system, which works well for low noise images. The proposed method is proven computationally efficient and sufficiently accurate for use in card games such as Baccarat.

The intention of this project is to build a playing card recognition system as an aid for a playing card game. The system has the capability to recognize a standard deck of playing cards, with both ranks and suits uniquely recognized. Recognizing playing cards involves character segmentation, affine transformation, edge detection and template matching. Rotation and scaling are considered for robust playing card recognition. The accuracy of card rank recognition with proposed method is about 99.79 percent in low noise circumstance, and the accuracy of suit recognition is about 81.06 percent.

This is a single pass playing card recognition system but is closest to the system proposed by this paper but with a lower accuracy of detecting suits.

## 3. Implementation

Matlab 7.1 was used to implement the system. Only functions that are available with Matlab 7.1 are used. The processing flow for the object detection pass is shown in Figure 1. The templates used to identify the playing card objects are in **Error! Reference source not found.**. The flow chart for determining suits and ranks is shown in **Error! Reference source not found.**. The templates used to find suits and ranks are shown in Figure **3**. They are extracted from the same cards that are to be compared and converted to binary images.

The system is implemented in two passes. Pass 1 is the object recognition of the playing cards where the image is first captured, and then it goes to template determination, classification and comparison, followed by playing card selection. In this step, the program differentiates the playing card from other objects using geometric selection. After scaling and rotating the cards, and comparing with two templates of the left and right edges, the playing cards are detected by using a simple subtractive image comparison with the stored template.

For the output, the face up playing cards will be bounded with green borders while other rectangle objects will be bounded with blue borders and non-rectangular objects will be bounded with red borders.

The semantic analysis where the suits and ranks of cards is determined, is done at another pass. That pass assumes that the cards had been identified and rotated to a vertical orientation. The top-left portion of the card containing both rank and suit is compared with a template of ranks and suits, using normalised cross correlation.

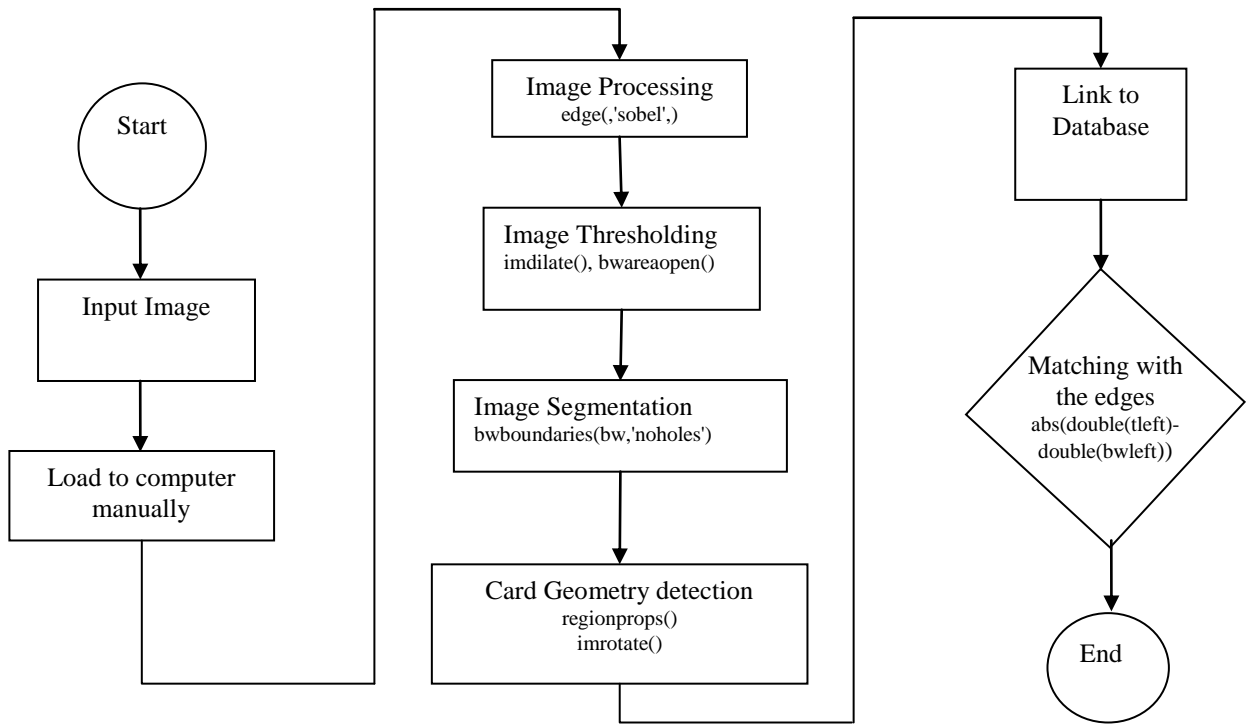

Figure 1: Object detection pass

The steps can be described as converting to greyscale, gaussian filtering, histogram equalization, edge detection, image convolution, morphological closing, removal of small area, cropping of bounded area, flood filling, computing cross-correlation with templates, finding maximum cross-correlation and labelling input according to maximum correlation.

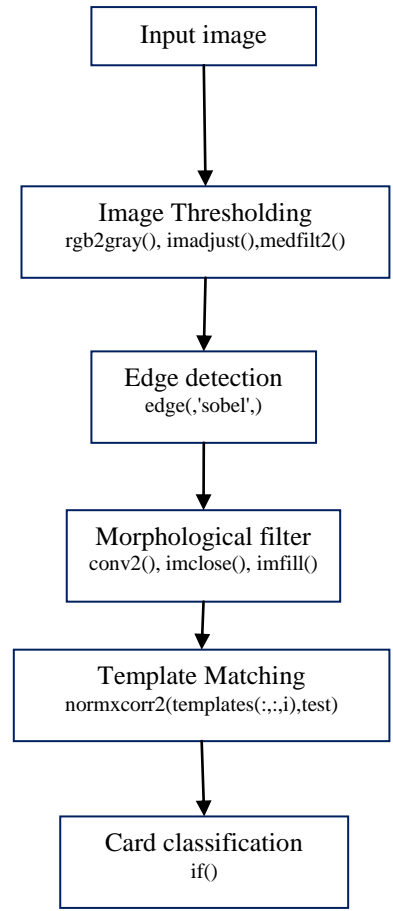

Figure 4: Semantic analysis pass

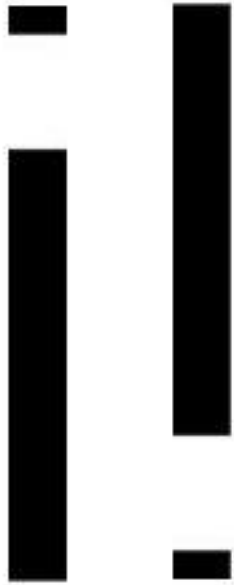

Figure 2: Left and right card edges templates

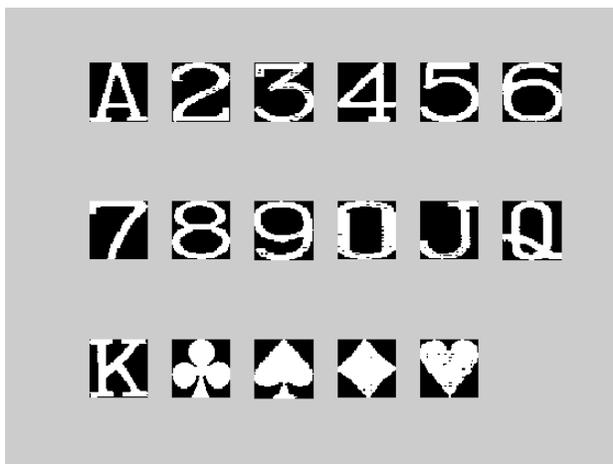

Figure 3: Templates to find suits and ranks

# 4. Results

## 4.1. Object detection pass

Objects that are bounded with green lines indicate objects that had been detected as playing cards as shown in Figure **5**.

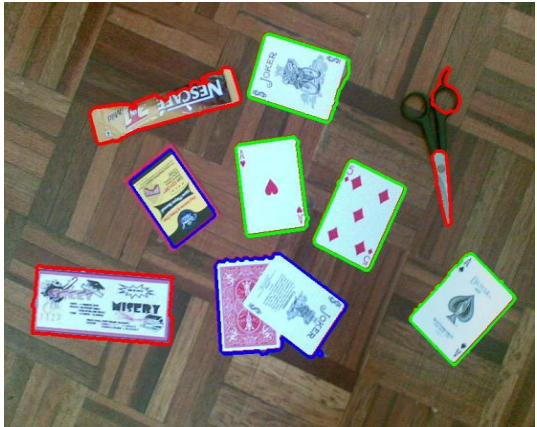

Figure 5: Detectable case

## 4.2. Object detection failures

Figure **6** shows a failure case. This is for a case of overlapping playing cards.

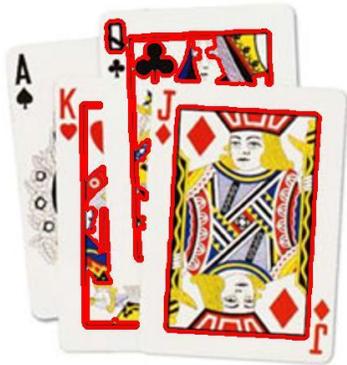

Figure 6: Failure cases

The highest percentage of testing accuracy was 16 successful detections out of 18 objects.

## 4.3. Semantic analysis pass

An experiment had been carried out to test the effectiveness of the purposely written cross-correlation template matching. In the experiment, the playing card recognition system is used to recognise the 52 different card images in vertical orientations. Every playing card was photographed 10 times to test the effectiveness of the semantic analysis. Therefore, each of those 52 playing cards was tested 10 times with different photos of similar rank and suit. The best 52 selected photos of the 52 playing card were also tested at reduced image sizes of 90%, 80%, 70%, 60%, 50%, 40%, 30% and 20%.

Overall detection rate for spade suit = 99.23%
Overall detection rate for heart suit = 100%
Overall detection rate for club suit = 99.23%
Overall detection rate for diamond suit = 96.92%

The poor detection rate for the diamond suit is due to the poor quality of pictures used. The diamond suit pictures were over exposed.

## 4.4. Diamond suit at reduced sizes

| Ratio / Rank | 1.0 | 0.9 | 0.8 | 0.7 | 0.6 | 0.5 | 0.4 | 0.3 |
|---|---|---|---|---|---|---|---|---|
| A | Yes | Yes | Yes | Yes | Yes | Yes | Yes | Yes |
| 2 | Yes | Yes | Yes | Yes | Yes | Yes | Yes | Yes |
| 3 | Yes | Yes | Yes | Yes | Yes | Yes | Yes | Yes |
| 4 | Yes | Yes | Yes | Yes | Yes | Yes | Yes | No |
| 5 | Yes | Yes | Yes | Yes | Yes | Yes | No | No |
| 6 | Yes | Yes | Yes | Yes | Yes | Yes | Yes | No |
| 7 | Yes | Yes | Yes | Yes | Yes | Yes | Yes | Yes |
| 8 | Yes | Yes | Yes | Yes | Yes | Yes | Yes | Yes |
| 9 | Yes | Yes | Yes | Yes | Yes | Yes | Yes | Yes |
| 10 | Yes | Yes | Yes | Yes | Yes | Yes | No | Yes |
| J | Yes | Yes | Yes | Yes | Yes | Yes | Yes | Yes |
| Q | Yes | Yes | Yes | Yes | Yes | Yes | Yes | Yes |
| K | Yes | Yes | Yes | Yes | Yes | Yes | Yes | No |

Table 1: Detection ability at reduced sizes

In Table **1**, the entry with a " yes" means that detection is possible.

## 4.5. Image vs. templates

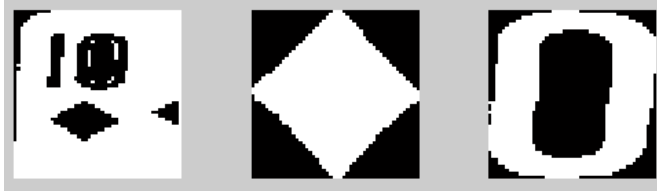

Figure 7: Diamond suit with rank 10

Figure 7 shows a picture, which is sampled from a Diamond suit with rank 10 on the left most part. On the right are the templates, which will be matched with the sampled image in order to recognise its rank and suit. This system only recognises single digits so the number ten is recognised as a card, which contains the number zero.

## 5. Conclusion and recommendations

A comparatively complete system of object detection and meaning extraction had been developed using Matlab 7.1 and shown to function within reasonable accuracy. Although the verification tests are not exhaustive, it should be clear that the results match closely the published detection results for the other published works in playing card detection when the test pictures are without much noise.

These noise free environments should be the default environments for most image recognition applications such as casinos or well lit car parks because humans need good lighting in order to function efficiently.

Matlab was shown to be able to develop such workable system at present without writing any extra functions. Although Matlab has functions for more sophisticated functions for image processing, even the standard normalised cross-correlation function has proven to be sufficiently good enough for the template matching routine. Semantic analysis, which is the extraction of meaning from the symbols, can be performed using Matlab's high-level language features such as the if-then-else statements.

By developing the whole system for detecting cards in casinos, it is clear that the main problem is in the card detection stage instead of the rank and suit determination.

The system had been done in two passes in order to improve performance. Rank and suit determination can be done for detected cards only instead of the whole image of the casino table.

Although it leads to problems of not evaluating undetected overlapping cards, the rank and suit determination can be done on non-playing cards that are rectangular objects, which are marked with blue borders.

To improve detection of overlapping cards, the segmentation can be extended to include holes as well. The number of templates for the edges had to be increased and template matching improved by using normalised cross correlation.

To improve the detection of ranks and suits, the templates can use greyscale instead of binary templates and samples of ranks and suits taken in various exposure levels.

All these improvements will use more computing resources that users may not have or afford. It is still useful to have a basic system that uses readily available resources in order to accomplish common problems.

Once resources become cheaper while faster, such as later versions of Matlab or similar software, performance and flexibility can be easily improved further. In the meantime, this research work shows what is readily available for the moment.

## Appendix: Internet Sources

## Appendix: Selected Matlab functions used

### Object detection:
EDGE(IM,'SOBEL', THRESHOLD * FUDGEFACTOR)
Find edges in intensity image.

IMDILATE(BW,SE1)  Dilation, enlarges foreground.

BWAREAOPEN(BW,NOISE) Binary area open; remove small objects.

BWBOUNDARIES(BW,'NOHOLES') Trace region boundaries in binary image.

REGIONPROPS(L,{'BOUNDINGBOX','PERIMETER','AREA','MAJORAXISLENGTH','MINORAXISLENGTH','ECCENTRICITY','ORIENTATION'})
Measure properties of image regions.

### Semantic analysis:

IM2BW() Convert image to binary image by thresholding.

MEDFILT2(A,[M N]) Performs median filtering of the matrix A in two dimensions.

EDGE(IM1,'SOBEL',THRESHOLD*FUDGEFACTOR)

CONV2() Two-dimensional convolution.

IMCLOSE() Close image.

IMFILL() Fill image regions.

NORMXCORR2(TEMPLATE,A)
Computes the normalized cross-correlation of matrices TEMPLATE and A.